\documentclass[letterpaper, 10 pt, conference]{ieeeconf}
\IEEEoverridecommandlockouts
\usepackage{graphicx}
\usepackage{subfigure}
\usepackage{hyperref}
\usepackage{mathrsfs}
\usepackage{amsmath,amssymb,amsfonts}
\usepackage[linesnumbered, ruled]{algorithm2e}
\usepackage{algorithmic}
\usepackage{multirow}
\usepackage{booktabs}
\usepackage{pifont}

\title{\LARGE \bf
GATER: Learning Grasp-Action-Target Embeddings and Relations for Task-Specific Grasping
}

\author{Ming Sun$^{1}$ and Yue Gao*$^{2}$
\thanks{$^{1}$Ming Sun is with Department of Automation, Shanghai Jiao Tong University, Shanghai, P.R. China. Email: {mingsun@sjtu.edu.cn}}
\thanks{$^{2}$Yue Gao is with MoE Key Lab of Artificial Intelligence and AI Institute of Shanghai Jiao Tong University, Shanghai, P.R. China. Email: {yuegao@sjtu.edu.cn}}
}

\begin{document}
\maketitle
\thispagestyle{empty}
\pagestyle{empty}

\begin{abstract}
Intelligent service robots require the ability to perform a variety of tasks in dynamic environments. Despite the significant progress in robotic grasping, it is still a challenge for robots to decide grasping position when given different tasks in unstructured real life environments. In order to overcome this challenge, creating a proper knowledge representation framework is the key. Unlike the previous work, in this paper, task is defined as a triplet including grasping tool, desired action and target object. Our proposed algorithm GATER (Grasp--Action--Target Embeddings and Relations) models the relationship among grasping tools--action--target objects in embedding space. To validate our method, a novel dataset is created for task-specific grasping. GATER is trained on the new dataset and achieve task-specific grasping inference with 94.6\% success rate. Finally, the effectiveness of GATER algorithm is tested on a real service robot platform. GATER algorithm has its potential in human behavior prediction and human-robot interaction. 
\end{abstract}

\section{Introduction}
Benefiting from the development of deep learning methods and the collection of large-scale datasets \cite{he2016deep, he2017mask,jiang2011efficient, mahler2017dex, depierre2018jacquard, levine2018learning}, many data-driven grasp detection methods have been proposed in recent years \cite{lenz2015deep, zeng2018robotic, chu2018real, mahler2019learning}. These methods achieved satisfactory success rate and generalization ability on novel objects for Pick-and-Place tasks. However, when considering task-specific grasping and execution of subsequent manipulation tasks, only optimizing grasp robustness is not enough. Therefore, depending on different tasks, the robot should perform different grasp configurations \cite{csahin2007afford, fang2020learning, ardon2019learning}.

Task-specific grasping methods aim to solve the above challenges \cite{bohg2013data, dang2014semantic, song2015task}. To find both stable and functionally suitable grasp allowing for the execution of a specific task, the robot requires the ability of reasoning considering the environmental context and the task requirements, which is more challenging than simply grasping an object \cite{song2015task, kokic2020learning}. In the past few years, many data-driven methods are proposed for leaning task-specific grasping, from manually labelled data, synthetic data, or human demonstrations \cite{fang2020learning, kokic2020learning, liu2020cage}. The experiment results demonstrate the effectiveness for task-specific grasping of a single object. For example, the predicted grasp position is located at the handle of the brush for cleaning, while located at the tip for handing over \cite{fang2020learning, dang2014semantic, kokic2020learning}.

\begin{figure}[t]
\centerline{\includegraphics[width=\columnwidth]{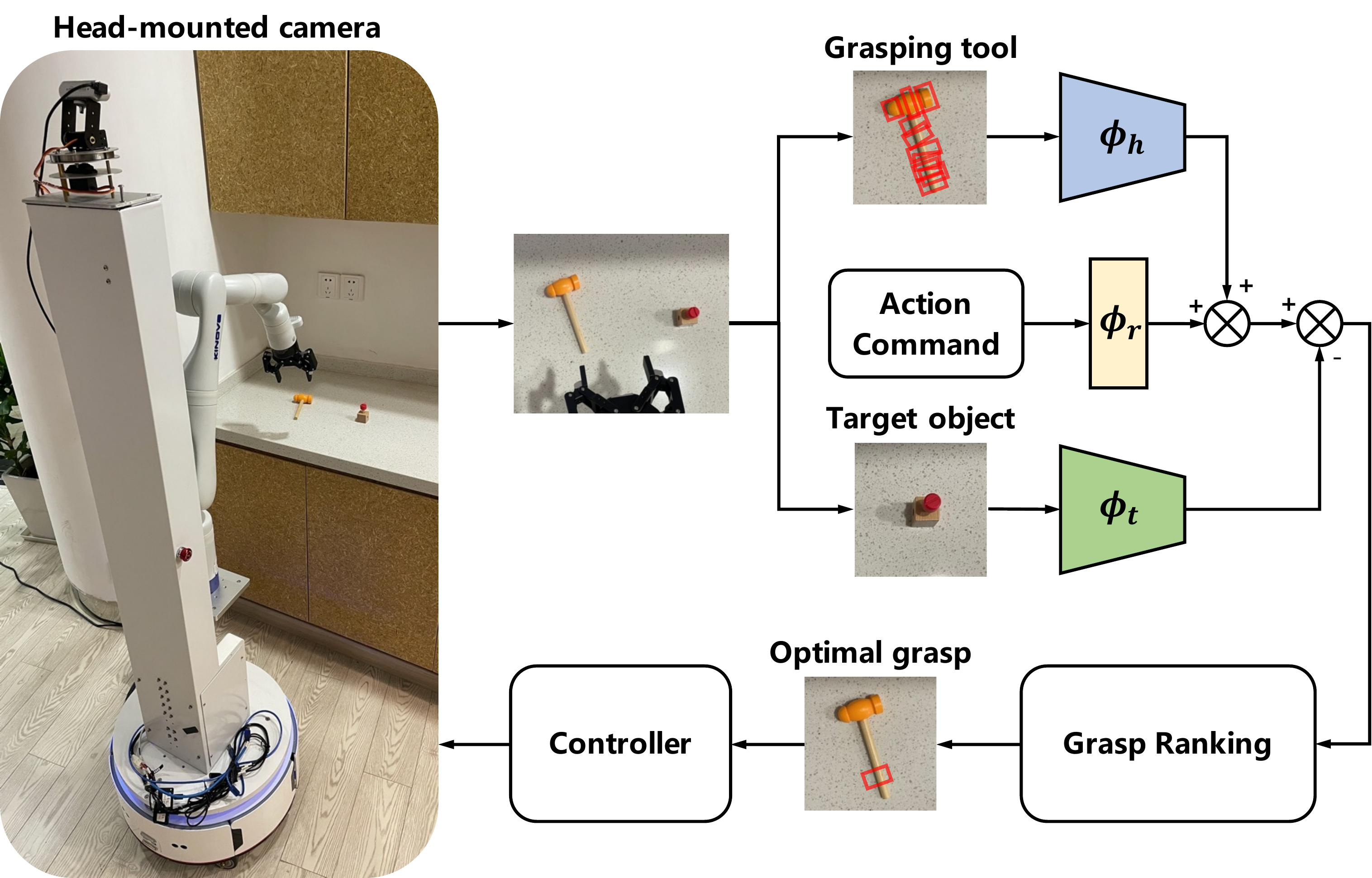}}
\caption{Illustration of task-specific grasping detection framework of GATER based on knowledge embedding. GATER extracts knowledge embeddings from RGB-D images of the objects and instruction of the desired action, evaluates the suitability of grasp candidates and determines the functionally optimal grasp configuration for the task.}
\label{schematic}
\end{figure}

Current task-specific grasp methods mostly formulate grasping task by considering the features of the object to be grasped and the desired action, while with little consideration of the target object \cite{saito2021select}. This is reasonable for binary actions such as hand-over, and pick-and-place tasks. Nevertheless, other actions such as knocking, cutting and cleaning, target object of the action is an essential part of the problem. It should be considered in the formulation of the task and should impact the computation of the selected tool and its grasp position. Imagine a service robot in the kitchen, it has a set of available tools, such as brush, duster, mop, sponge, steel wool, etc. Its task is to clean the dishes, for which it has to decide the grasp position of the tool. In this scenario, current methods will fail because all the tools contain the parts with the action ``clean". Therefore, no matter the task is to clean the dishes, shoes, floors or glasses, the robot may execute the same grasp configuration by choosing the highest probability of ``clean'' and ignoring the differences of the target objects.

From human's common sense, for task-specific grasping, the grasping tool, the desired action and the target object are all very important. Hence, the information of these three parts should be considered in task-specific grasping. In this work, we propose a knowledge embedding method GATER (Grasp--Action--Target Embeddings and Relations) for task-specific grasping. The task is defined as a triplet which includes grasping tool, desired action and target object, as shown in Fig. \ref{schematic}. Different from previous knowledge-embedding-based task-specific grasping methods \cite{liu2020cage, daruna2019robocse, murali2020same}, GATER determines task-specific grasping considering the grasping tool, the desired action and the target object. This approach has a wider definition for grasping tasks and is more intuitive just like human. In addition, GATER does not need a separate stage to extract abstract knowledge or semantics such as material or shape. Hence, the contributions of this work are as the following:

\begin{itemize}
\item GATER algorithm defines task-specific grasping with three components: the potential grasping tool, the desired action and the target object.
\item An end-to-end method is proposed to automatically learn the embeddings for task-specific grasping, which can directly learn the intrinsic attributes of the objects from the perception information and generalize to unseen tasks, without hand-designed semantics.
\item A task-specific grasping dataset containing grasping tools, desired actions and target objects with labelled annotations is created. GATER is trained on the dataset. In addition, the learned model is tested on a realistic service robot platform Kinova Gen3 to demonstrate the validity of our approach.
\end{itemize}

\section{Related Works}
\label{relatedworks}
\subsection{Task-Agnostic Grasping}
Classical geometry-based grasping methods such as force closure \cite{nguyen1988constructing} and form closure \cite{liu2000computing} highly rely on the known 3D object model and the precise physical information. Therefore, geometry-base grasping methods are difficult to apply to daily objects. Over the last two decades, data-driven methods have significantly advanced the progress on robotic grasping. Jiang et al. proposed a 7-dimensional grasping rectangle including gripper's location, orientation and opening width, and a SVM ranking algorithm was learned to give the proper grasp output \cite{jiang2011efficient}. Meanwhile, Cornell Grasping Dataset was collected, which later became a benchmark dataset and greatly promoted the progress of data-driven grasp detection methods \cite{lenz2015deep, chu2018real, redmon2015real, kumra2017robotic, morrison2018closing}. 

With the introduction of several more large-scale datasets, deep learning methods started to became popular for robotic grasping in recent years \cite{jiang2011efficient, mahler2017dex, depierre2018jacquard}. Lenz et al. employed a simplified 5-dimensional grasping rectangle representation \cite{lenz2015deep}, and applied a two-stage sliding window detection pipeline with two deep neural networks in detecting robotic grasps. Redmon et al. split the image into N$\times$N grid and performed single-stage grasp bounding boxes regression to avoid intensive computational cost and achieve real-time detection \cite{redmon2015real}. In \cite{chu2018real}, grasp detection was treated as a combination of regression and classification problems, and multiple grasp candidates for multiple objects can be predicted in a single shot. The light-weight GG-CNN predicts pixel-wise grasp quality in a depth image and outputs the best antipodal grasp, allowing for real-time detection up to 50Hz \cite{morrison2018closing}. 

The above works pioneered the introduction of deep learning methods into robotic grasping and significantly advanced the progress of this field. However, due to the fact that only optimizing grasp robustness without considering subsequent tasks, these methods may sometimes fail to select a functionally suitable grasp for the task \cite{fang2020learning}. In this work, we focus more on how to determine the optimal grasp configuration for a certain task when multiple grasp candidates are available. 

\subsection{Task-Specific Grasping}
Task specific grasping refers to object grasping which allows for the execution of a specific task \cite{bohg2013data, kokic2020learning}. Analytical methods solve the problem of task-specific grasping based on task wrench space \cite{li1988task, haschke2005task}. Recently, data-driven methods illustrate the advantage in the ability of reasoning and generalization. Dang et al. proposed an example-based framework to generate task-specific grasps using partial object geometry, tactile contacts and hand kinematic data \cite{dang2014semantic}. Song et al. \cite{song2015task} presented a probabilistic framework for the representation and the modeling of task-specific grasping based on Gaussian mixture models and discrete Bayesian networks. Another effective approach is to learning task-specific grasping from human demonstration and human-robot interaction \cite{kokic2020learning, balasubramanian2012physical}. 

In recent years, several related large-scale datasets are created \cite{myers2015affordance, do2018affordancenet, ardon2019learning, murali2020same}. Ard{\'o}n et al. utilized Markov Logic Networks to learn grasp affordance based on semantics \cite{ardon2019learning}. Unlike previous methods which are limited to a single hypothesis, this approach can detect multiple grasps for a single object with the consideration of environmental context such as location. Similarly, Murali et al. proposed GCNGrasp to learn relations between objects and tasks with Graph Convolutional Network \cite{murali2020same}.

Currently, task-specific grasping considering target objects are attracting some research attention. Task-Oriented Grasping Network (TOG-Net) jointly optimizes both grasping and manipulation policy for tools \cite{fang2020learning}. The authors predefined two tasks, which include the grasping tool, the desired action and the target object. Saito et al. agreed that the robot must both understand grasping tool and target object for grasping, and constructed a deep neural networks to recognize the characteristics of target objects \cite{saito2021select}. In this paper, the definition of task is similar to \cite{fang2020learning} and \cite{saito2021select}, but GATER explicitly defines the three elements of a task and learns knowledge embeddings for them to model grasp--action--target relationship.

\subsection{Knowledge Embedding for Grasping}
In robotics, knowledge embedding methods can be utilized to assist the understanding of the environment around the robots. Many prior methods, such as \cite{zhu2014reasoning}, KnowRob \cite{tenorth2013knowrob}, RoboCSE \cite{daruna2019robocse}, attempt to create a general knowledge engine for robots to learn knowledge representations and carry out various tasks. Some recent works focus on knowledge embedding for more specific tasks, such as grasping. Jang et al. proposed an object-centric visual embedding based on object persistence for robotic manipulation tasks from self-supervised learning \cite{jang2018grasp2vec}. CAGE presents a semantic representation of grasp contexts to inform grasp selection \cite{liu2020cage}. GCNGrasp utilizes a Graph Convolutional Network to build a knowledge graph in which each node is encoded as a D-dimensional embedding \cite{murali2020same}. In GATER, a translational distance model, such as TransE \cite{bordes2013translating}, TransH \cite{wang2014knowledge}, TransR \cite{lin2015learning}, is utilized to learn knowledge embeddings for grasping tools, desired actions and target objects. 

\begin{figure}[t]
\centering
\subfigure[]{\label{definition1}\includegraphics[width=4.0cm]{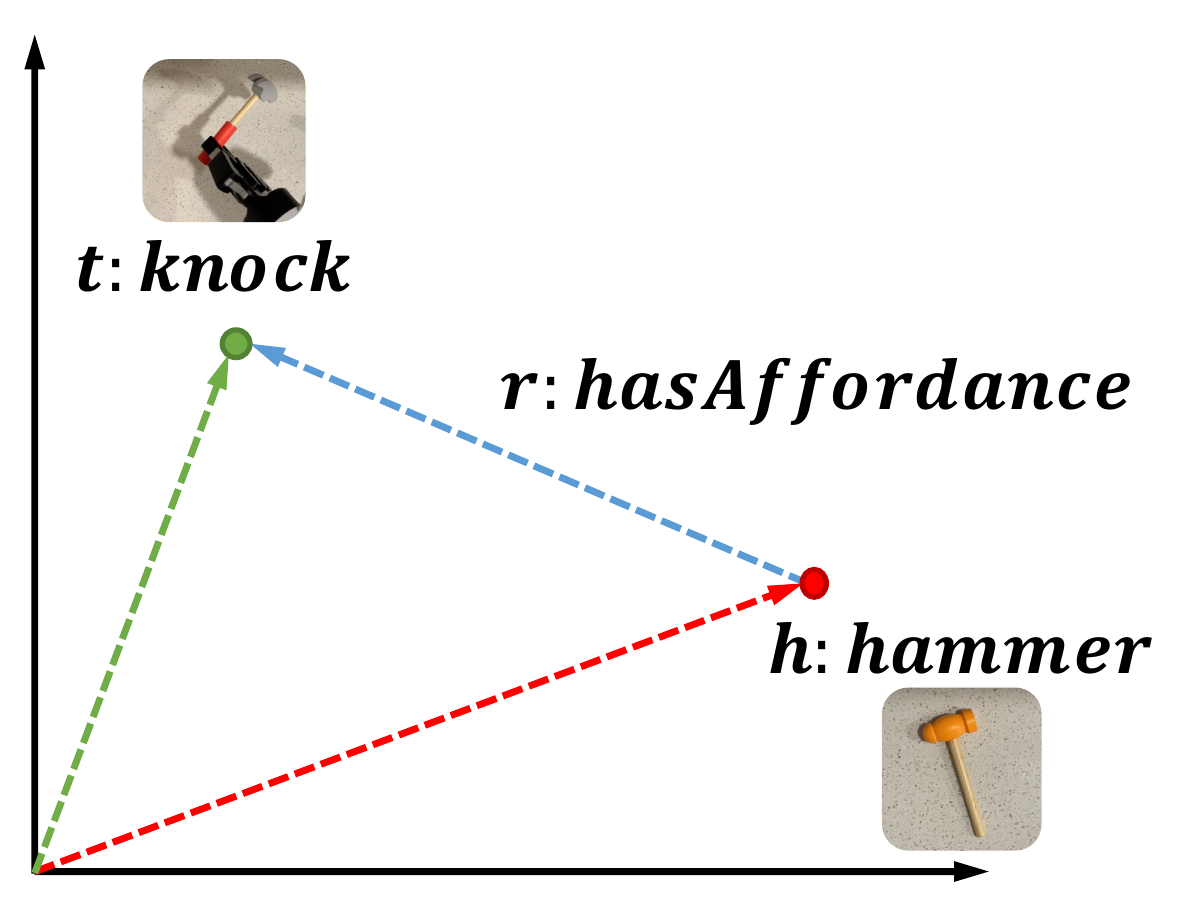}}
\subfigure[]{\label{definition2}\includegraphics[width=4.0cm]{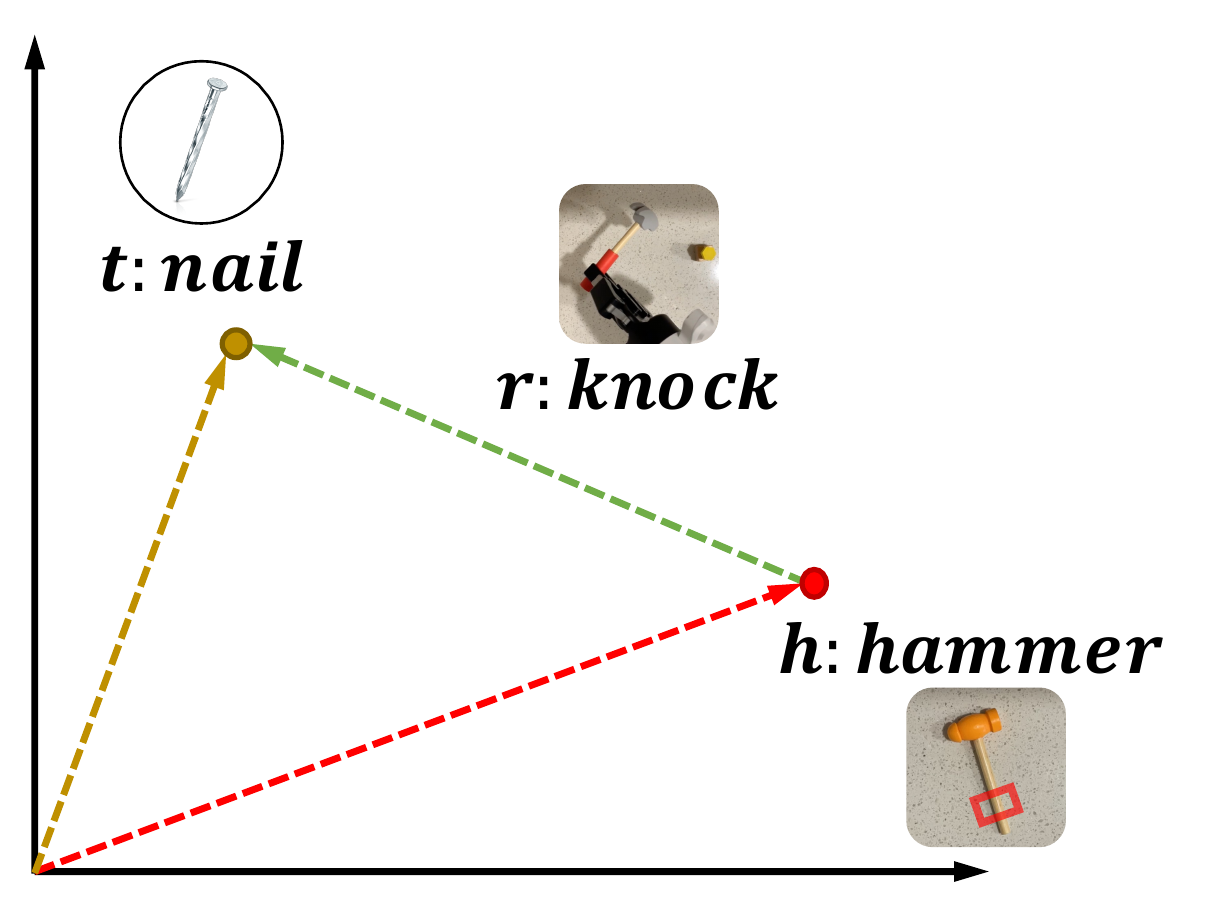}}
\caption{Comparison of task definitions between (a) previous works \cite{daruna2019robocse, zhu2014reasoning, tenorth2013knowrob}, and (b) GATER. In GATER, the grasping tool, desired action and target object are encoded as $h$, $r$ and $t$, respectively.}
\label{definition}
\end{figure}

\section{Problem Statement}
\label{problem}
In this paper, we consider the problem of selecting the most suitable grasp for a task given a set of stable grasp candidates from any reliable grasp detection method. The visual perception of the objects and the desired action command are provided for robot to understand task and generate grasp candidates, which are then ranked by their suitability for the task.

\vspace{0.1cm}

\noindent \textbf{Task Definition:} Inspired by translation-based models \cite{bordes2013translating, wang2014knowledge, lin2015learning}, we consider the triplet in the form of $(head, relation, tail)$ (denoted as $(h, r, t)$). Previous works usually define the grasping tool and the desired action as $h$ and $t$ of the triplet without considering the target object, such as $(hammer, hasAffordance, knock)$ shown in Fig. \ref{definition1} \cite{daruna2019robocse, zhu2014reasoning, tenorth2013knowrob}. In GATER, the grasping tool and the target object are defined as $h$ and $t$, and the desired action is defined as $r$, such as ($hammer, knock, nail$), as shown in Fig. \ref{definition2}. This definition is more reasonable since objects are encoded as entities and actions are encoded as relations. Hence the relation $r$ can express more specific meaning. Furthermore, the binary actions can be converted to the triplets by generalizing the target object as the entity $None$. 

\vspace{0.1cm}

\noindent \textbf{Notations:} For the task triplet defined above, let $\mathcal{O}=(\mathcal{O}_h,\mathcal{O}_t)$ be the visual observation space, which respectively denote the grasping tool and the target object. Let $\mathcal{A}$ denote the action space and $N$ denote the size of action, where each action $a \in \mathcal{A}$ is represented by a one-hot encoding. For a given grasping observation $o_h \in \mathcal{O}_h$, a set of grasp candidates $\mathcal{G}$ are available.

In our framework, a task is defined as $T=(o_h, a, o_t)$. GATER learns grasp--action--target embeddings and relations for the task through
\begin{equation}
\phi(T)=(h,r,t)=(\phi_h(o_g), \phi_r(a), \phi_t(o_t))
\end{equation}
where $o_g$ is the grasping tool with a grasp $g \in \mathcal{G}$, and $\phi$ is the knowledge embedding function.

\vspace{0.1cm}

\noindent \textbf{Objective:} Given the observation $o \in \mathcal{O}$ and the desired action $a \in \mathcal{A}$ with a set of grasp candidates $\mathcal{G}$, the goal is to measure the suitability of $g \in \mathcal{G}$ for the task $T=(o_h, a, o_t)$ and select the optimal grasp configuration $g^*$ from $\mathcal{G}$ based on the grasp suitability for the specific task: 
\begin{equation}
\label{objective}
g^*=\mathop{\arg\max}_{g \in \mathcal{G}} Q(g|o,a)
\end{equation}

\begin{figure*}[t]
\centerline{\includegraphics[width=16.9cm]{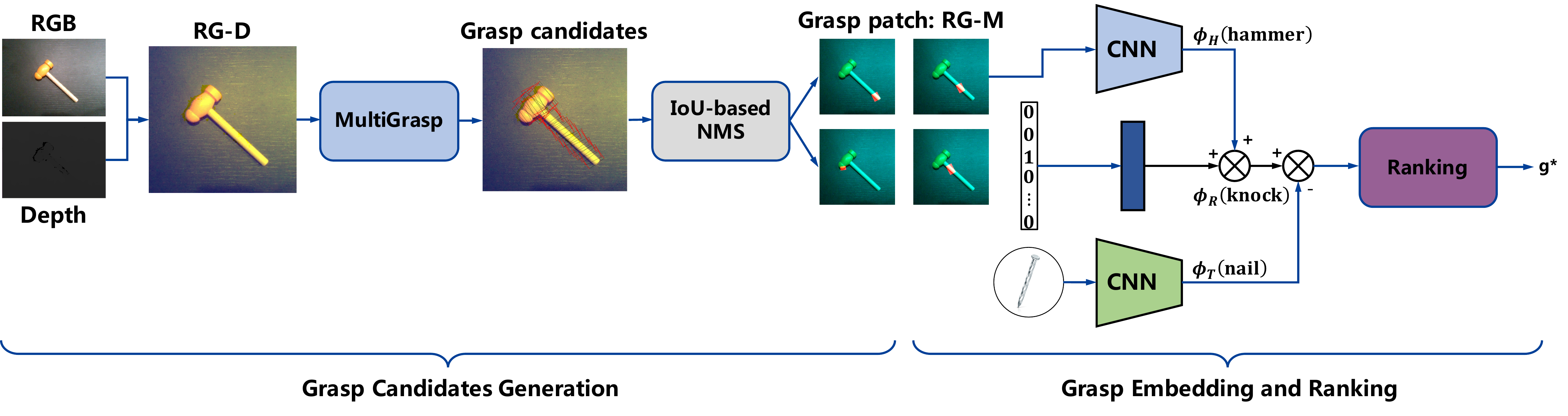}}
\caption{Framework of GATER for task-specific grasping, comprised of a grasp candidates generation module and a grasp embedding and ranking module. The model is trained in an end-to-end manner and predicts the functionally optimal grasp based on the suitability for the given task. The RG-M data is rendered as M-GR for clarity in figure presentation.}
\label{overview}
\end{figure*}

\section{Methods}
In order to address the problem under the task definition in Section \ref{problem}, we propose GATER, which takes visual observation $o$ and desired action $a$ as input and outputs the most suitable grasp for the task. As shown in Fig. \ref{overview}, GATER is comprised of a grasp candidates generation module and a grasp embedding and ranking module. These two modules are cascaded and trained in an end-to-end manner. In this section, we first introduce the process of grasp candidates generation and the conversion to RG-M format for grasping tools, and describe the design of the model architecture. Then we present the end-to-end learning method with the triplet training set and the inference procedure for task-specific grasping using the trained model.

\subsection{Grasp Candidates Generation}
As introduced in Section \ref{relatedworks}, task-agnostic grasp detection has been extensively studied and achieved satisfactory performance. Hence the multi-grasp detection approach presented in \cite{chu2018real} is adopted as the grasp candidates generator, where the input is the RG-D image of the object. The detected grasps are represented as a 5-dimensional grasp rectangle
\begin{equation}
\label{grasp}
g=\{x, y, \theta, w, h\}^T
\end{equation}
where $(x, y)$ is the center of the grasp, $\theta$ is the orientation of the grasp, $h$ is the opening distance of the gripper and $w$ is the width of the gripper. 

The multi-grasp detection approach usually outputs a large number of grasp candidates and many grasp rectangles with large overlapping regions may reside in the same part of an object. Hence to ensure that grasp regions are uniformly distributed in different parts of the object, Non-Maximum Suppression is employed on a small subset of grasp candidates $\mathcal{G}_\alpha \subseteq \mathcal{G}$ with grasp quality higher than $\alpha$. Inspired by \cite{morrison2018closing}, each remaining grasping rectangle is converted to a grasp mask. Then the depth channel in RG-D image is substituted into grasp mask to obtain RG-M image of the grasping tool, which represents an object with a grasp for learning knowledge embedding.

\subsection{Network Architecture for Knowledge Embedding}
For grasp embedding and ranking module, it takes as input the RG-M image $o_g$ of the grasping tool generated by the above-mentioned grasp candidates generation module, the RGB image of the target object $o_t$ and the one-hot encoding of the desired action $a$, and outputs the ranking of the grasps ordered by their suitability for the task based on the embeddings. In GATER, three separate neural networks $\phi_h$, $\phi_r$ and $\phi_t$ are employed for learning embeddings from $o_g$, $a$, $o_t$ to $h$, $r$, $t$, respectively. Two independently computable and trainable ResNet-50 $\phi_h$ and $\phi_t$ work as feature extractors for $h$ and $t$ respectively, and the output of the last residual block is taken as the embedding of the input image. A two-layer fully-connected network $\phi_r$ is used for learning embeddings for $r$, where all hidden layers have 2048 units and the dimension of the output is consistent with $h$ and $t$ to satisfy the additivity. The parameters of the three networks are initialized randomly and updated from learning. 

\subsection{Learning with End-to-End Training}
GATER learns vector embeddings of the entities and the relationships from training set $\mathcal{T}$ of triplets $(o_g, a, o_t)$, representing that the object with grasp patch in the image $o_g$ can apply action $a$ to the target object in $o_t$, such as $(hammer, knock, nail)$. For binary actions without target objects, a generalized entity $None$, expressed as a blank image, is utilized to denote that $o_t$ is not required. Given a training triplet, we want $h+r=t$ holds, i.e., $h+r$ should be nearest to $t$, while be far away from $t$ otherwise, where $h$, $r$, $t$ are respectively the embeddings of $o_g$, $a$, $o_t$. To learn such embeddings, the loss function introduced in \cite{bordes2013translating} is adopted as 
\begin{equation}
\label{lossaff}
\mathcal{L}_{aff}=\sum_{(h, r, l) \in S} \sum_{(h', r, l') \in S'_{(h, r, l)}}[\gamma+d(h+r, t)-d(h'+r, t')]_+
\end{equation}
where $\gamma>0$ is the margin hyper-parameter, $d$ is the $L_1$-norm to measure similarity, and
\begin{subequations}
\label{embedding}
\begin{align}
h &= \phi_h (o_g)    \label{eqa}\\
r &= \phi_r (a)      \label{eqb}\\
t &= \phi_t (o_t)    \label{eqc}
\end{align}
\end{subequations}
\vspace{-0.5cm}
\begin{equation}
\label{setS}
S=\{(h,r,t) | \exists (o_g, a, o_t) \in \mathcal{T} \ \ satisfy \ \ Eq.\ (\ref{embedding}) \}
\end{equation}

Except modeling relationships, the consistency should also be satisfied when learning embeddings. The embedding consistency means that the entities with similar characteristics or attributes in different images should be embedded into vectors that are closer to each other. To this end, two cross-entropy loss functions $\mathcal{L}_{h, cls}$ and $\mathcal{L}_{t, cls}$ are defined for classification of $o_g$ and $o_t$. The feature extractors for embedding and classification in each ResNet-50 share an identical set of residual blocks to guarantee embedding consistency, and both $\mathcal{L}_{h, cls}$ and $\mathcal{L}_{t, cls}$ can integrate multiple classification problems to provide the model with more prior information to facilitate learning knowledge embeddings, such as object category, grasp region, and semantic attributes defined in previous works like shape, texture and material. Hence, GATER is trained to learn knowledge embeddings by optimizing the following loss function
\begin{equation}
\label{loss}
\mathcal{L}=\mathcal{L}_{aff}+\mathcal{L}_{h, cls}+\mathcal{L}_{t, cls}
\end{equation}
One of the benefits of above procedure is that the model can be trained in an end-to-end manner, without the need for the extra semantics extractors.

\subsection{Inference for Task-Specific Grasp}
Algorithm \ref{algorithm} describes the outline of end-to-end task-specific grasp prediction procedure with GATER, which aims to determine the grasp configuration from visual perception for the task. Given visual observation $o$ and desired action $a$, grasp candidates generation module will provide a set of reliable grasp candidates, which are then filtered by grasp quality and Non-Maximum Suppression, and converted to a small set of $o_g$ in RG-M format. Then the learned model is used to extract knowledge embeddings of $a$, $o_t$ and each $o_g \in \mathcal{O}_g$, and calculate the grasp suitability of each $o_g$ for the task, which is measured by the inverse of $L_1$-norm $d(h+r-t)$. Thus the grasps in $\mathcal{O}_g$ can be ranked by their suitability for the task. The optimal grasp $g^*$ is transformed from image space to 3D world coordinates and sent to the robot and executed. 

Compared with previous works, the learning and the inference of GATER is more like to model the corresponding relationship between the task and the attributes of the object with grasps, e.g., for knocking a nail, a hard tool with a long handle and a large contact surface should be used. GATER can reason about the desired attributes of the object with a grasp rather than one specific object, and then select the best match. This approach based on knowledge embedding is more consistent with human's understanding for tasks and reasoning for grasp selection and is closer to learning the intrinsic attributes of the objects, instead of directly predicting a grasp only considering the grasping object. 

Another difference between our approach and previous methods is that, instead of learning a uni-directional mapping from objects to actions, GATER learns the relational embeddings of objects and actions, which allows multi-directional prediction from given scenario. In this way, GATER may also have the potential ability on human behavior prediction and human-robot interaction, which can help the robot to determine the grasp according to the commands from humans and assist the robot to predict the intentions of humans.

\begin{algorithm}[t]
\caption{GATER Prediction}
\label{algorithm}
\textbf{Input:} Camera perception $o=(o_h,o_t) \in \mathcal{O}$\;
\textbf{Input:} desired action $a \in \mathcal{A}$\;
Grasp candidate set $\mathcal{G} \leftarrow$ $multiGrasp(o_h)$\;
Subset of grasp set with high quality grasps $\mathcal{G}_\alpha \subseteq \mathcal{G}$\;
Subset of grasp set $\mathcal{G}_{\alpha}' \leftarrow$ $NMS(\mathcal{G}_\alpha)$\;
RG-M with grasp mask $\mathcal{O}_g \leftarrow$ $rectToMask(\mathcal{G}_{\alpha}')$\;
\For{$o_g \in \mathcal{O}_g$}
{
Knowledge embedding: $h=\phi_h(o_g)$, $r=\phi_r(a)$, $t=\phi_t(o_t)$\;
Grasp suitability prediction for the task: Calculate $Q(g|o,a)=\frac{1}{d(h+r-t)}$\;
}
Find $g^*=\mathop{\arg\max}_{g \in \mathcal{G}} Q(g|o,a)$\;
Transform grasp coordinate $g^*$ in image space into 3D world coordinate $g_r=t_{RC}(t_{CI}(g))$\;
\textbf{Output:} Send $g_r$ to robot to execute the grasp.
\end{algorithm}

\section{Experiments}
To the best of our knowledge, all current grasping datasets lack the consideration of targets objects. Therefore, to experimentally validate our approach, a novel dataset for task-specific grasping is collected and manually annotated. Then GATER is trained by end-to-end learning on the dataset. The learned model is deployed in real-world grasping scenarios on a mobile manipulator platform equipped with a lifting platform and a 7-DoF Kinova Gen3 robotic arm with a parallel-jaw gripper Robotiq 2F-85 and a RGB-D camera Intel RealSense D435 mounted on top of the platform. 

\subsection{Assumptions}
We assume the observation $o$ are captured with a fixed overhead RGB-D camera with know intrinsics and precise calibration between the camera and the robot base. Based on these assumptions, a grasp in the image space $g$ can be converted to a grasp in world coordinate $g_R$ through Eq. (\ref{convert}) and then be sent to the robot to execute, where $t_{CI}$ converts the grasp $g$ from image coordinate to the 3D coordinate of camera frame, and $t_{RC}$ transforms from camera frame to robot frame. 
\begin{equation}
\label{convert}
g_R=t_{RC}(t_{CI}(g))
\end{equation}

\subsection{Dataset}
\begin{figure}[t]
\centerline{\includegraphics[width=8.5cm]{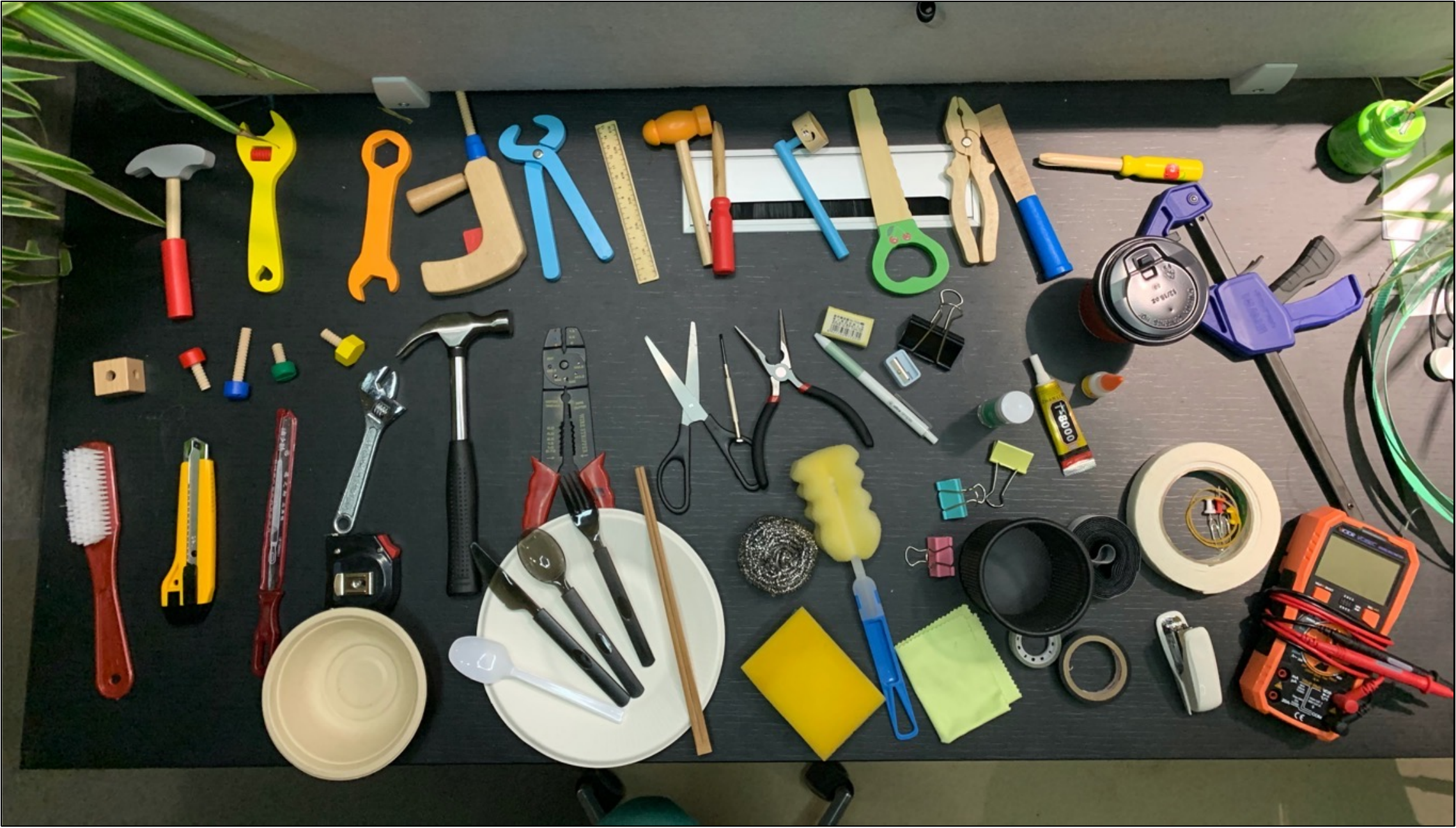}}
\caption{Dataset collection. The dataset contains 56 grasping tools, 27 target objects and 10 desired actions.}
\label{dataset}
\end{figure}

In order to establish a dataset for task-specific grasping under the definition described in Section \ref{problem}, we collect the RGB and the depth images of grasping tools and target objects, and annotate the desired actions, as shown in Fig. \ref{dataset}. In particular, this dataset contains 56 grasping tools with 206 images and 1,079 grasp regions, 27 target objects with 98 images, 10 actions, and 12,505 hand-labelled triplets in the form of $(o_g, a, o_t)$. The triplets in the dataset are split into 70\% training and 30\% testing in image-wise split and object-wise split. 

\subsection{Qualitative Analysis}
\label{qualitativeanalysis}
\begin{figure}[t]
\centering
\subfigure[]{\label{qualitative1}\includegraphics[width=\columnwidth]{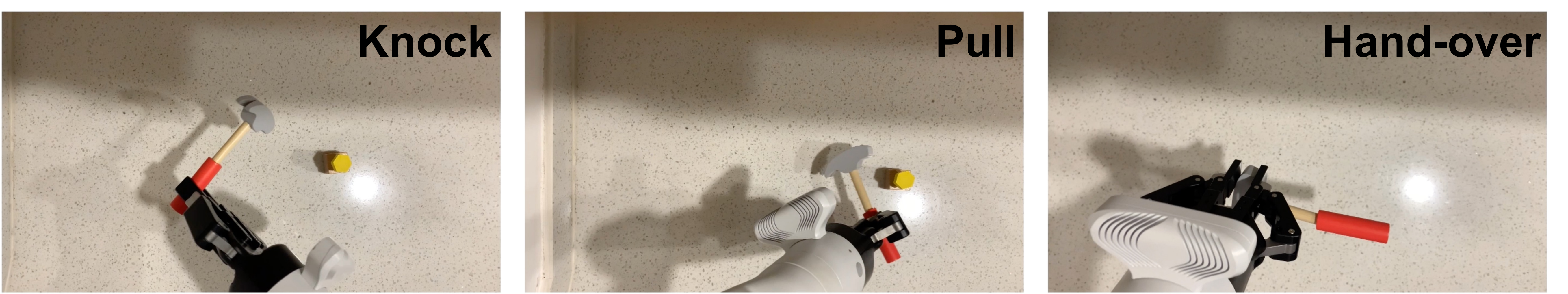}}\\
\subfigure[]{\label{qualitative2}\includegraphics[width=\columnwidth]{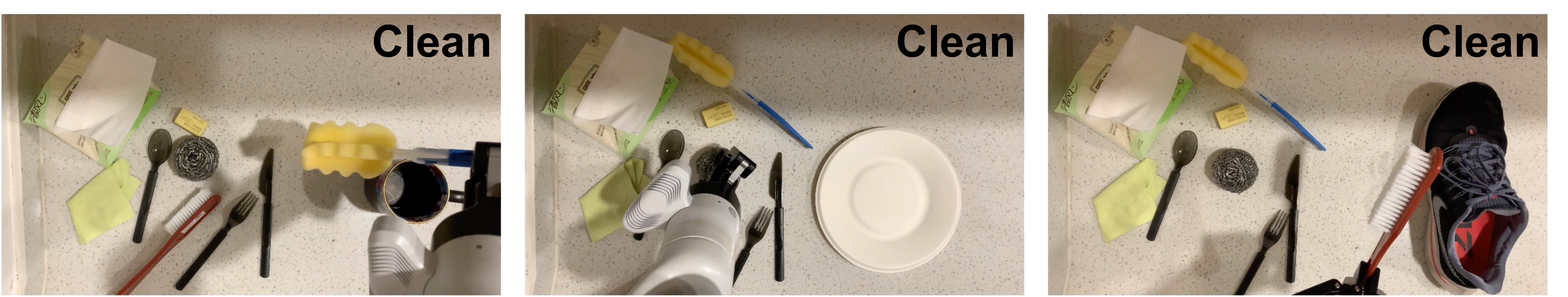}}\\
\subfigure[]{\label{qualitative3}\includegraphics[width=\columnwidth]{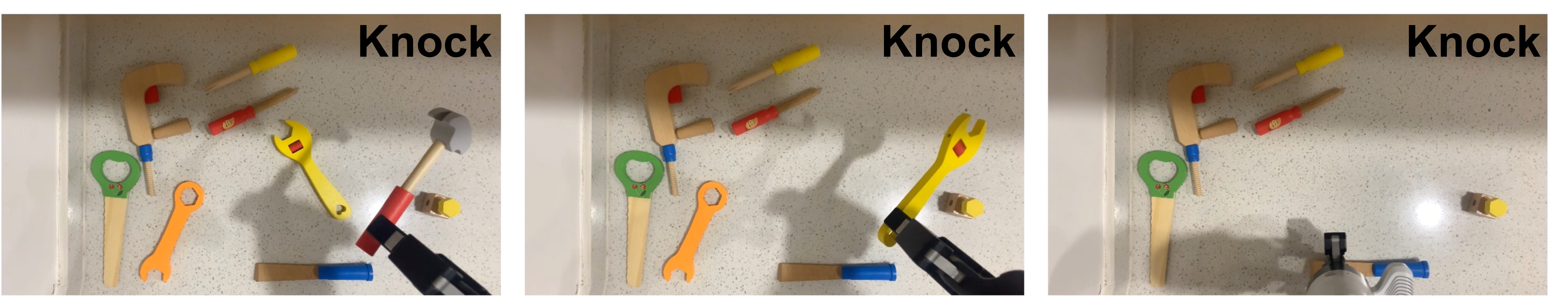}}
\caption{Qualitative experiment results in Section \ref{qualitativeanalysis}. (a) Select different grasps for different desired actions. (b) Select different tools with grasp for different target objects. (c) Select the sub-optimal option for the task based on the embedding of the desired grasp when the tool set changes.}
\label{qualitative}
\end{figure}

In Fig. \ref{qualitative}, we show the qualitative experiment results on realistic robotic arm to demonstrate that GATER can select different grasps for different tasks when any element in $(o_h, a, o_t)$ changes. 

Fig. \ref{qualitative1} shows that for the same grasping tool and target object, GATER can select different grasps for different desired actions. For the desired action $knock$ and $pull$, the robot tends to grasp the far end of the handle to lift up the hammer and use the hammer head to perform actions on the nail. And for $hand-over$, the robot learns to grasp the hammer head and hand the handle to others. 

For different target objects, GATER can select suitable grasps according to the task. As shown in Fig. \ref{qualitative2}, for the same desired action $clean$ and the same set of grasping tools, GATER leans to select sponge brush for mug, steel wool for plate, and scrubbing brush for shoes. Previous methods without considering target objects cannot handle these tasks well because all grasping tools in this set contain parts related to the action ``clean". Therefore, for different target objects, these method always output the same grasp with highest probability of ``clean" because target objects are not taken into account in the model. 

Another benefit of the end-to-end knowledge embedding learning is that in GATER, the learned embedding represents the desired attributes of the object with a grasp, rather than a specific entity. Therefore, when the optimal grasp is not available, GATER can select a sub-optimal option. For instance, to knock a nail in Fig. \ref{qualitative3}, the embedding of the desired grasp may be ``grasp the handle of a hard tool with a long handle and a large contact surface". The most suitable option for the tool set is to grasp the handle of the hammer. When the hammer is removed from the set, GATER can grasp the handle of the wrench according to the similarity of the embedding, which may not have the direct relation to the action ``knock" but in current tool set is the closest to the desired grasping tool. Further, if the wrench is also removed, it will grasp the head of the file and use the handle to knock the nail. This procedure can be viewed as reasoning about the grasp based on the understanding of tasks and the attributes of objects, which can assist the robot to better understand and perform tasks in changeable environments. 

\begin{figure}[t]
\centering
\subfigure[]{\label{embeddingsa}\includegraphics[width=4.2cm]{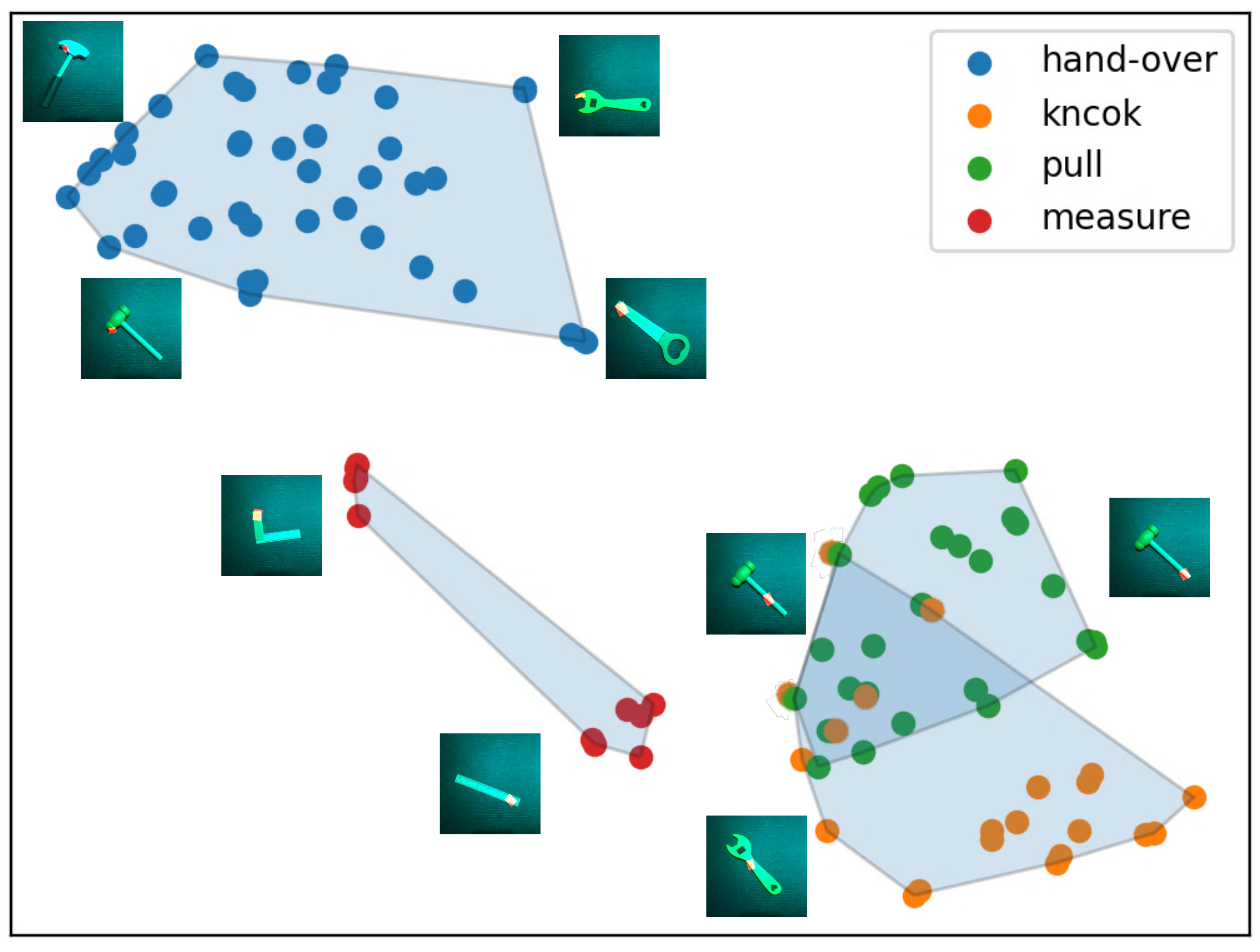}}
\subfigure[]{\label{embeddingsb}\includegraphics[width=4.2cm]{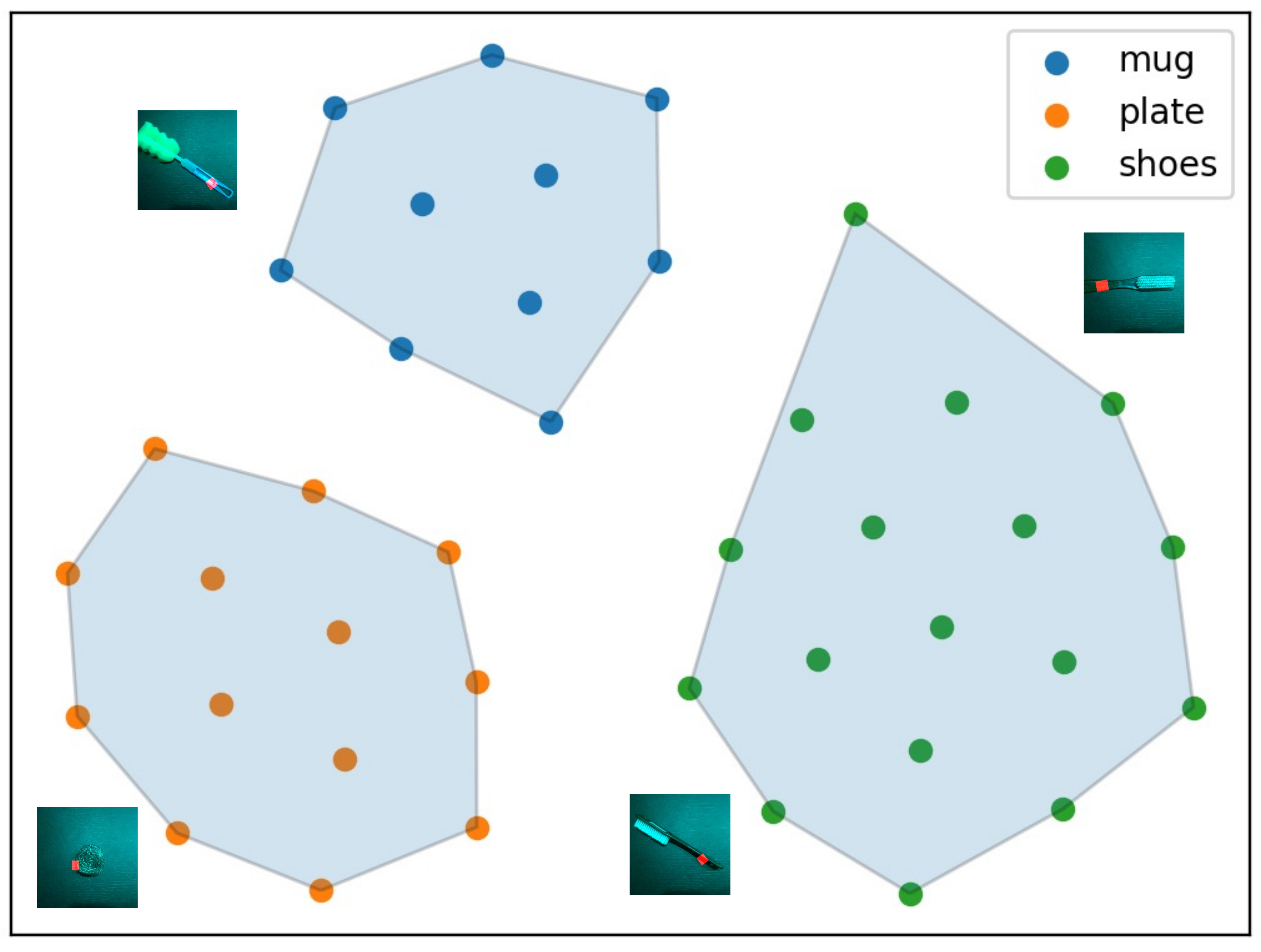}}
\caption{t-SNE results on the embedding space for dimensionality reduction: (a) Embeddings of grasping tools for fixed target object $nail$ and different desired actions. (b) Embeddings of grasping tools for fixed desired action $clean$ and different target objects.}
\label{embeddings}
\end{figure}

Fig. \ref{embeddings} shows the results of applying t-SNE algorithm to the embedding space for dimensionality reduction, with different desired actions and different target objects. In Fig. \ref{embeddingsa}, the target object is fixed as a nail for knocking, pulling and measuring. The different colors in the plot indicate the embeddings of grasping tools for different actions ``hand-over'', ``knock'', ``pull'' and ``meausre''. As shown in the figure, the embeddings of grasping tools for ``knock'' and ``pull'' are closer to each other with intersection as GATER decides to grasp the handle of tools for both ``knock'' and ``pull''. Fig. \ref{embeddingsb} shows the embeddings of the grasping tools for different target objects for action ``clean'', where GATER decides to grasp steel wool for cleaning plate, sponge for mug and brush for shoes. The experiment results show that the embeddings of grasping tools with similar attributes and functions are clustered, which indicate that GATER can model the relationship among grasping tool--desired action--target object in embedding space to decide functionally suitable grasp for task-specific grasping.

\subsection{Quantitative Evaluation}
A high-quality task-specific grasp should simultaneously satisfy: (\romannumeral1) the robustness to stably hold the object and (\romannumeral2) the suitability to the planning and the execution for subsequent manipulation tasks. For this reason, the evaluation metric includes two parts -- task-agnostic metric (grasp robustness) and task-specific metric (task suitability). For task-agnostic quality evaluation,  grasping detection method introduced in \cite{chu2018real} is utilized as the baseline. Following the evaluation metric in previous work \cite{chu2018real, redmon2015real, kumra2017robotic}, a predicted grasp candidate $g^*$ is correct for task-agnostic metric if: (1) The difference between the angle of $g^*$ and a ground truth grasp configuration is less than $30^{\circ}$, and (2) The Jaccard index of $g^*$ and $30^{\circ}$ is greater than 0.25. For task-specific metric, a grasp candidate is reported as correct for the task if the predicted grasp $g^*$ satisfies the above criterion with the ground truth grasp $g$, where the triplet $(o_g, a, o_t)$ is correct for the task.

Compared with prior works, the definition of task in GATER has substantially changed. To our knowledge, previous datasets and approaches lack the consideration of target objects, which are summarized in Table \ref{comparison}. In this case, the comparison between GATER and previous methods is actually unfair as they do not consider the information of target objects. Therefore, the comparison experiments are conducted in two manners, as shown in TABLE \ref{comparisonResult}. For target-agnostic grasping, the target-agnostic variant of GATER is trained and compared with \cite{do2018affordancenet} and \cite{daruna2019robocse}, where the target objects are fixed to ``None'' for GATER. For target-known grasping, the information of target object is available, but the approaches in \cite{daruna2019robocse, do2018affordancenet} and target-agnostic GATER do not consider target object. 

To further investigate the effect of desired actions and target objects, an ablation study is performed on our approach. Specifically, we compare the default model with a model without grasp embedding and ranking module and a model with all target objects set to ``None". The quantitative evaluation results of the three models lead to Table \ref{ablation}. GATER performs slightly worse on task-agnostic metric. This is reasonable because except for grasp robustness, GATER also needs to consider task-specific metric, i.e., the trade-off between grasp robustness and task suitability to obtain a grasp that is acceptable for both robustness and suitability. Therefore, GATER cannot always select the most robust and stable grasp. For task-specific metric, since GATER takes complete task representation into account, it outperforms all competitive methods, which demonstrates the advantages of GATER's task representation and model. 

Table \ref{physical} shows the results of GATER on real robot platform. In this experiment, 10 unseen tasks in testing set are selected and 50 trials are performed per task for a total of 500 evaluations. The ``Scene 2'' in the table indicates a set of grasping tools in Fig. \ref{qualitative2}. The prediction is considered as correct if both task-agnostic and task-specific metrics are satisfied, and the grasping success indicates the grasping is stably completed on real robot. 

\begin{table}[]
\scriptsize
\centering
\caption{Comparison Between Prior and Our Datasets and Methods}
\label{comparison}
\begin{tabular}{cccccccccc}
\toprule
Dataset & Grasping Tool & Desired Action & Target Object \\
\midrule
Cornell \cite{jiang2011efficient} & \ding{52} & \ding{56} & \ding{56}\\
Dex-Net 2.0 \cite{mahler2017dex} & \ding{52} & \ding{56} & \ding{56}\\
Jacquard \cite{depierre2018jacquard} & \ding{52} & \ding{56} & \ding{56}\\
RGB-D Part \cite{myers2015affordance} & \ding{52} & \ding{52} & \ding{56}\\
IIT-AFF \cite{do2018affordancenet} & \ding{52} & \ding{52} & \ding{56}\\
Aff-Reasoning \cite{ardon2019learning} & \ding{52} & \ding{52} & \ding{56}\\
TaskGrasp \cite{murali2020same} & \ding{52} & \ding{52} & \ding{56}\\
Ours & \ding{52} & \ding{52} & \ding{52}\\
\midrule
Approach & Grasping Tool & Desired Action & Target Object \\
\midrule
\cite{lenz2015deep} & \ding{52} & \ding{56} & \ding{56}\\
\cite{redmon2015real} & \ding{52} & \ding{56} & \ding{56}\\
\cite{kumra2017robotic} & \ding{52} & \ding{56} & \ding{56}\\
\cite{chu2018real} & \ding{52} & \ding{56} & \ding{56}\\
\cite{do2018affordancenet} & \ding{52} & \ding{52} & \ding{56}\\
\cite{daruna2019robocse} & \ding{52} & \ding{52} & \ding{56}\\
\cite{ardon2019learning} & \ding{52} & \ding{52} & \ding{56}\\
\cite{murali2020same} & \ding{52} & \ding{52} & \ding{56}\\
\cite{fang2020learning} & \ding{52} & \ding{52} & \ding{52}\\
\cite{saito2021select} & \ding{52} & \ding{52} & \ding{52}\\
Ours & \ding{52} & \ding{52} & \ding{52}\\
\bottomrule
\end{tabular}
\end{table}

\begin{table}[]
\scriptsize
\centering
\caption{Results of Comparison Between Existing Methods and GATER}
\label{comparisonResult}
\begin{tabular}{cccccccccc}
\toprule
Approach & Target-agnostic grasping & Target-known grasping\\
\midrule
AffordanceNet \cite{do2018affordancenet} & 87.2\% & 77.1\%\\
RoboCSE \cite{daruna2019robocse} & 82.6\% & 72.8\%\\
Target-agnostic GATER & 80.5\% & 76.4\%\\
GATER & -- & 85.1\%\\
\bottomrule
\end{tabular}
\end{table}

\section{Conclusions}
\subsection{Discussions}
This work addresses the problem of task-specific grasping, where the task is explicitly defined as a triplet including grasping tool, desired action and target object. We introduce GATER, an end-to-end knowledge embedding learning approach based on translation-based method to model the relationship among these three elements of a task. A task-specific grasping dataset is collected under this task definition. Using this novel dataset, GATER can learn knowledge embeddings for task-specific grasping and infer grasp position for tasks. Experiment results show that the grasp selected by GATER is more suitable than previous methods for various tasks. 

In addition to above benefits, we would like to discuss the inherent characteristics and advantages of GATER:

\textbf{(\romannumeral1) Mapping v.s. Embedding:} Most previous works try to establish a mapping from object image to task or a logic inference model based on semantics, while GATER aims to model the relationship among grasping tool, desired action and target object. We would like to mention that the embedding of GATER is not for a specific entity, but for the attributes of the desired object with a suitable grasp. For example, for a task triplet $(\underline{~~~}, knock, nail)$, the embedding of the blank we expect should express something like ``grasp the handle of a hard tool with a long handle and a large contact surface", rather than a specific tool ``hammer". According to this embedding, the robot can determine the best grasping tool with the grasp position based on the ranking of suitability, such as $hammer > wrench > pliers$. This relation modeling method based on knowledge embedding is more consistent with humans on the understanding and the reasoning for tasks.

\begin{table}[]
\scriptsize
\centering
\caption{Ablation Stduy}
\label{ablation}
\begin{tabular}{cccccccccc}
\toprule
\multirow{2}{*}{Model} & \multicolumn{2}{c}{Image-wise split} & \multicolumn{2}{c}{Object-wise split} \\
\cmidrule(r){2-3} \cmidrule(r){4-5}
& Task-agnostic & Task-specific
& Task-agnostic & Task-specific\\
& metric & metric & metric & metric\\
\midrule
$o_h$        & 95.5\% & 87.1\% & 95.1\% & 74.9\%\\
$o_h + a$    & 94.6\% & 90.5\% & 94.4\% & 76.4\%\\
GATER    & 94.7\% & 94.6\% & 94.2\% & 85.1\%\\
\bottomrule
\end{tabular}
\end{table}

\begin{table}[]
\scriptsize
\centering
\caption{Experimental Evaluation on Real Robot Platform}
\label{physical}
\begin{tabular}{ccc|ccccccc}
\toprule
Grasping & Desired & Target & Prediction & Grasping & Success \\
Tool & Action & Object & Success & Success & Rate\\
\midrule
Hammer & Hand-over & -- & 46/50 & 42/50 & 84\%\\
Hammer & Knock & Nail & 43/50 & 43/50 & 86\%\\
Hammer & Pull & Nail & 44/50 & 42/50 & 84\%\\
Bowl & Contain & Rice & 46/50 & 42/50 & 84\%\\
Spoon & Scoop & Rice & 41/50 & 32/50 & 64\%\\
Ruler & Measure & Nail & 45/50 & 24/50 & 48\%\\
Scissors & Cut & Tape & 40/50 & 31/50 & 62\%\\
Scene 2 & Clean & Mug & 40/50 & 39/50 & 78\%\\
Scene 2 & Clean & Plate & 42/50 & 39/50 & 78\%\\
Scene 2 & Clean & Shoes & 44/50 & 42/50 & 84\%\\
\midrule
& Total& & 431/500 & 376/500 & 75.2\%\\
\bottomrule
\end{tabular}
\end{table}

\textbf{(\romannumeral2) Uni-directional v.s. Multi-directional prediction:} Most of previous methods can only perform uni-directional reasoning from object to action, while GATER can infer the unknown term from any other two given terms in $(h, r, t)$ based on the translational distance of the embeddings. This property can empower robots on human behavior prediction and human-robot interaction. For instance, when a robot sees a person holding a hammer and a nail, it might predict that the person may want to knock the nail. When it sees a boy grasping the handle of a mug, it should infer that he may want to drink and then pour some water for him; or in another scenario where a girl is holding the rim of a mug, the robot should speculate that she may want to hand over the mug, hence it should try to reach out and take over the mug. 

\subsection{Limitations and Future Works}
At present, our method aimed at task-specific grasping, but the manipulation planning for task is not included. In addition, taking target object into consideration may not be sufficient enough for completing task-specific manipulation. The discussion and the exploration of what information should be considered for a task and how to encode these information worth further study and discussion.

In future works, we will further integrate trajectory planning into our framework to form a more complete system for task-specific grasping and manipulation based on embedding. This may involve the problem of how to encode other available information of the task. Moreover, we would like to explore the ability of GATER in human behavior prediction and human-robot interaction.

\bibliographystyle{IEEEtran}
\bibliography{IEEEabrv,mybibfile}

\end{document}